\documentclass{ecai}
\usepackage{times}
\usepackage{graphicx}
\usepackage{latexsym}

\usepackage{soul}
\usepackage{url}
\usepackage[hidelinks]{hyperref}
\usepackage[utf8]{inputenc}
\usepackage[justification=centering]{caption}
\DeclareCaptionLabelFormat{onepica}{#1 #2.\hspace{12pt}}
\usepackage{algorithm} 
\usepackage{algorithmic} 

\usepackage{bm} 

\usepackage{amsmath,amssymb,amsfonts}
\usepackage{booktabs}
\usepackage{algorithm}
\usepackage{algorithmic}
\usepackage{makecell}
\usepackage[inline]{enumitem}
\usepackage{subcaption}
\usepackage{mwe}
\usepackage{diagbox}
\usepackage{multirow}
\begin{document}

\title{BOASF: A Unified Framework for Speeding up \\ Automatic Machine Learning via Adaptive Successive Filtering}

\author{Guanghui Zhu \and Xin Fang \and Feng Cheng \and Lei Wang \\ \and Wenzhong Chen  \and Chunfeng Yuan \and Yihua Huang\institute{State Key Laboratory for Novel Software Technology, Nanjing University, China, email:\{xinfang,chengfeng,wizcheu\}@smail.nju.edu.cn, \{zgh,cfyuan,yhuang\}@nju.edu.cn, wanglwig@gmail.com}
}

\maketitle
\bibliographystyle{ecai}

\begin{abstract}
Machine learning has been making great success in many application areas. 
However, for the non-expert practitioners, it is always very challenging to address a machine learning task successfully and efficiently. 
Finding the optimal machine learning model or the hyperparameter combination set from a large number of possible alternatives usually requires considerable expert knowledge and experience. 
To tackle this problem, we propose a combined Bayesian Optimization and Adaptive Successive Filtering algorithm (BOASF) under a unified multi-armed bandit framework to automate the model selection or the hyperparameter optimization.
Specifically, BOASF consists of multiple evaluation rounds in each of which we select promising configurations for each arm using the Bayesian optimization. 
Then,  ASF can early discard the poor-performed arms adaptively using a Gaussian UCB-based probabilistic model. 
Furthermore, a \emph{Softmax} model is employed to adaptively allocate available resources for each promising arm that advances to the next round. 
The arm with a higher probability of advancing will be allocated more resources. 
Experimental results show that BOASF is effective for speeding up the model selection and hyperparameter optimization processes while achieving robust and better prediction performance than the existing state-of-the-art automatic machine learning methods.
Moreover, BOASF achieves better anytime performance under various time budgets.
\end{abstract}

\section{INTRODUCTION}
The great success of machine learning in many real-world areas has attracted more practitioners to solve their problems using machine learning algorithms. 
Due to the ever-increasing number of machine learning models, selecting the best model for the specific problem domain is often a tedious endeavor.
Moreover, even if the machine learning model has been determined, the hyperparameter search space is also very large. 
As a result, both the model selection and hyperparameter optimization processes require considerable expert knowledge and experience with the given problem domain. 
It is a challenging and time-consuming task for non-experts to find the best model or hyperparameter combination from a large number of possible alternatives. 
Therefore, the automatic machine learning (AutoML) algorithm that can choose good models or hyperparameters without any human intervention is becoming increasingly important.

In practice, the available resources for solving the AutoML problem is usually limited, especially in the application areas that require a good machine learning model in a short time.
Thus, given fixed resources (e.g., the total available time), a practical AutoML system should be able to fast converge to the best model or hyperparameter combination.
To achieve the goal, we need to consider the following problems. 
First, during the model selection process, the poor-performed models should be identified and discarded as early as possible.
Second, the promising models retained for further evaluation should be allocated more resources. 
Furthermore, the better-performed models should be given more opportunities to be evaluated.
Recently, the successive halving method proposed in Hyperband~\cite{DBLP:journals/jmlr/LiJDRT17} directly discards half of the candidate arms in a brute-force manner, making it possible to miss the potentially promising arms.

On the other hand, a practical AutoML system should support both model selection and hyperparameter optimization under the same algorithm framework. 
Unfortunately, most of the current AutoML methods usually focus on only one aspect of AutoML, either model selection or hyperparameter optimization. 
The existing AutoML tools such as AutoSklearn view the AutoML problem as a combined model selection and hyperparameter optimization problem~\cite{DBLP:conf/nips/FeurerKESBH15,DBLP:conf/kdd/ThorntonHHL13}. 
However, given a large number of available models, the combined hyperparameter space is very high-dimensional, which results in low search efficiency and slow convergence rate.

In this paper, we propose a combined Bayesian Optimization and adaptive successive filtering algorithm called BOASF to speed up the process of AutoML under a unified multi-armed bandit framework. 
Specifically, the overall process is divided into multiple evaluation rounds.
In each round, each arm will be evaluated multiple times with different configurations under given resources. 
Bayesian optimization is used to select promising configurations for each arm. 
Then, ASF computes the Gaussian Upper Confidence Bound (UCB) for each arm according to the mean and variance of the historical evaluation results. 
Next, the Gaussian UCB score is transformed to the probability of advancing to the next evaluation round. 
Furthermore, to adaptively determine the resources allocated to each promising arm, we employ a \emph{Softmax} model, where the arm with a higher probability of advancing will be allocated more resources. 
The selection process is repeated until the total resources run out. 
Finally, we can get the best evaluation result through the overall evaluation process. 

Moreover, BOASF can be easily extended to support both the model selection and the hyperparameter optimization. 
For the model selection, each available model is viewed as an arm. 
For the hyperparameter optimization, we propose a novel sub-space partitioning method to divide the hyperparameter search space into several disjoint sub-spaces. 
Each sub-space is regarded as an arm of BOASF. 
Thus, BOASF can handle model selection and hyperparameter automatically under the same algorithm framework. 
Moreover, BOASF is naturally apt to be parallelized.

Experimental results on a wide range of classification datasets show that BOASF is effective for speeding up both model selection and hyperparameter optimization while achieving robust and better prediction performance than the existing state-of-the-art AutoML methods. 
In the model selection scenario, BOASF statistically outperforms AutoSklearn in 23/30 cases and ties in 3 cases. 
Moreover, for hyperparameter optimization, BOASF achieves better results than just searching over the entire space in 23/30 cases.
Additionally, BOASF achieves better anytime performance under different time limits.

\section{PRELIMINARIES}
\subsection{The AutoML Problem}
Given a new problem domain, AutoML aims to search the best machine learning model and hyperparameter combination automatically. 
Correspondingly, the AutoML problem can further be divided into two subproblems: model selection and hyperparameter optimization.
\subsubsection{Model Selection}
Given a training set $D_{\rm{train}}$ and the available algorithm set $\mathcal{A}$, the goal of model selection is to determine the best-performed algorithm $A^*$ using $k$-fold cross-validation. Suppose that $D_{\rm{train}}$ is split into $K$ folds $\{D^{1}_{\rm{train}},\cdots,D^{K}_{\rm{train}}\}$ and $\{D^{1}_{\rm{val}},\cdots,D^{K}_{\rm{val}}\}$ such that $D^{i}_{\rm{train}}=D_{\rm{train}} \backslash D^{i}_{\rm{val}}$ for $i =1,\cdots,K$.
Let $\mathcal{L}(A, D^{i}_{\rm{train}}, D^i_{\rm{val}})$ be the loss that algorithm $A$ achieves on $D^{i}_{\rm{val}}$ when trained on $D^{i}_{\rm{train}}$ .
Formally, the model selection problem can be written as:
\begin{equation}
    A^{*} \in \arg \min _{A \in \mathcal{A}} \frac{1}{K} \sum_{i=1}^{K} \mathcal{L}\left(A, D^{i}_{\rm{train}}, D^{i}_{\rm{val}}\right).
\end{equation}

\subsubsection{Hyperparameter Optimization}
Given a learning algorithm $A$, the hyperparameters of $A$ also plays an important role in the final prediction performance. 
Suppose that algorithm $A$ contains $n$ hyperparameters $\lambda_1,\cdots,\lambda_n$ with domains $\Lambda_1,\cdots,\Lambda_n$, the hyperparameter space $\Lambda$ is the cross-product of these domains: $\Lambda = \Lambda_1 \times\cdots \times \Lambda_n$.
Given such a hyperparameter space $\Lambda$, we define the hyperparameter optimization problem as follows:
\begin{equation}
    \lambda^{*} \in \arg \min _{\lambda \in \Lambda}  \frac{1}{K} \sum_{i=1}^{K} \mathcal{L}\left(A, \lambda, D^{i}_{\rm{train}}, D_{\rm{val}}^{i}\right).
\end{equation}

\subsubsection{$\emph{\textbf{k}}$ Expensive Bandit Problem}
To address the model selection and hyperparameter optimization under the same framework, we introduce the $k$ expensive bandit problem ~\cite{DBLP:journals/jmlr/LiJDRT17,DBLP:journals/corr/abs-1012-2599}.
For example, each candidate model can be viewed as an arm of the bandit. 
Every arm $a$ will return a reward $r(a)$ after being evaluated. 
The reward can be the prediction accuracy of the learning model.
In practice, the cost of evaluating an arm is expensive. 
Therefore, the optimization objective of the $k$ expensive bandit problem is to search the arm with the optimal reward in the context of limited resources.
Let $R$ denote the total available resources, $B$ the arm set of the bandit, and $\Psi(a)$ the evaluation cost of the arm $a$. 
Formally, the objective of the $k$ expensive bandits problem is as follows:
\begin{equation}
    \begin{array}
        {l}{\arg \max \limits_{a \in B} r(a)} \\[4mm]
        {s.t. \sum \limits_{a \in B} \Psi(a) \leq R}.
    \end{array}
\end{equation}
Next, we introduce our proposed method that automates the model selection and hyperparameter optimization under a unified $k$ expensive multi-armed bandit framework. 


\section{THE PROPOSED METHOD}

In this section, we introduce our practical AutoML method to address model selection and hyperparameter optimization automatically.
The method is called BOASF since it combines Bayesian optimization (BO)
and adaptive successive filtering (ASF). 
BOASF is a non-stochastic finite $k$-armed strategy that adaptively allocates available resources among multiple arms.
For example, we view each candidate machine learning model as an arm. 
The naive strategy for $k$ expensive bandit problem is to allocate the equal resources to each arm and evaluate each arm independently. 
However, this equal-allocation method is inefficient in the context of model selection or hyperparameter optimization. 
This is because the performance may differ a lot between arms in practice. 
Some arms may perform much worse than others. 
Thus, we need to filter the arms with poor performance and allocate more resources for the potentially promising arms. 
To achieve the goal, we first propose an adaptive successive filtering approach, which
achieves both arm filtering and resource allocation adaptively according to the performance of each arm.

\begin{figure}
	\centering
	\includegraphics[width=0.99\columnwidth]{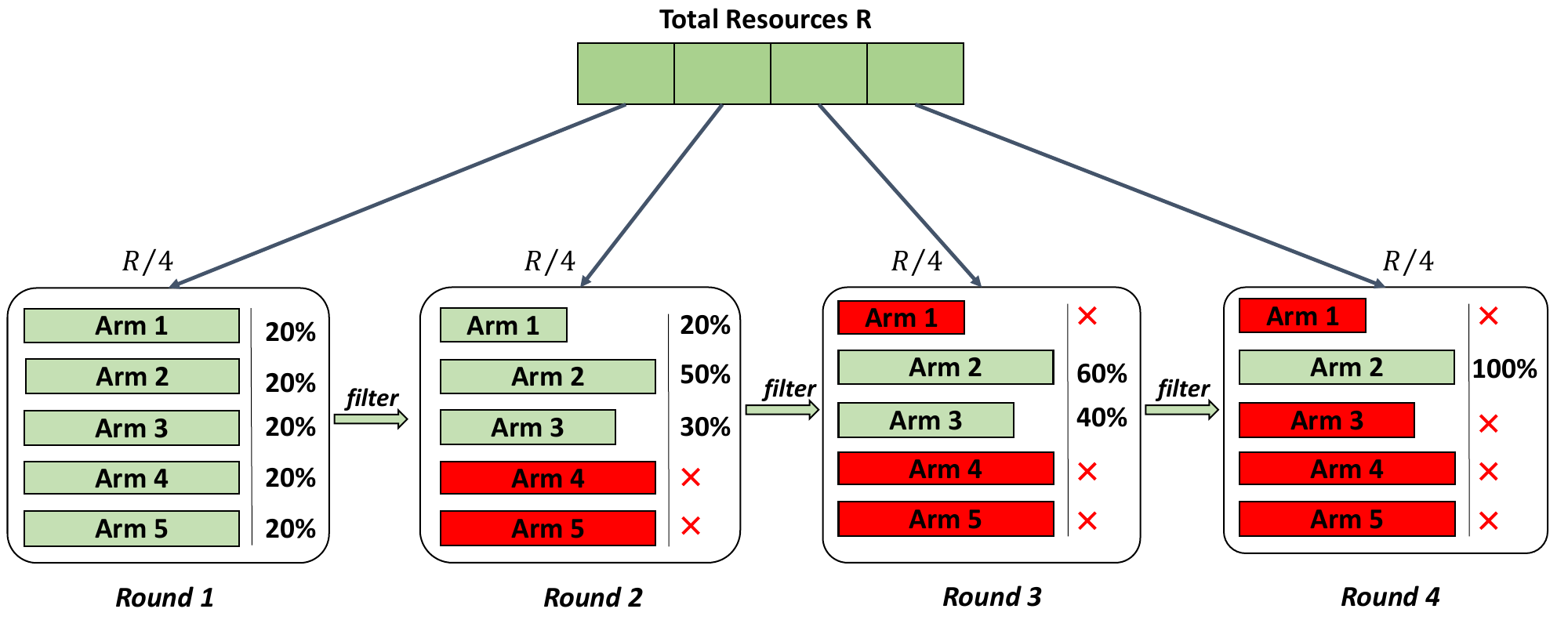}
	\caption{Overall workflow of a 4-round ASF}
	\label{fig:overflow}
\end{figure}

Figure~\ref{fig:overflow} shows the overall workflow of ASF, which is divided into multiple rounds.
The available resources in each round are equal.
Let $R$ denote the total available resources and $r$ the total rounds.
The resources allocated in each round is $R/r$.
In the first round, since the performance of each arm is unknown, we adopt the pure exploration strategy and allocate the resources equally.
Then, the arms are filtered progressively.
In each round, ASF evaluates each arm under the given resources.
Based on the evaluation results, poor-performed bandits will be discarded adaptively.
Meanwhile, a \emph{Softmax} model is employed to adaptively determine the resource allocation for each selected arm that advances to the next evaluation round.
To enhance exploitation, the arms with better performance will be allocated more resources.

\begin{figure}
	\centering
	\includegraphics[width=0.98\columnwidth]{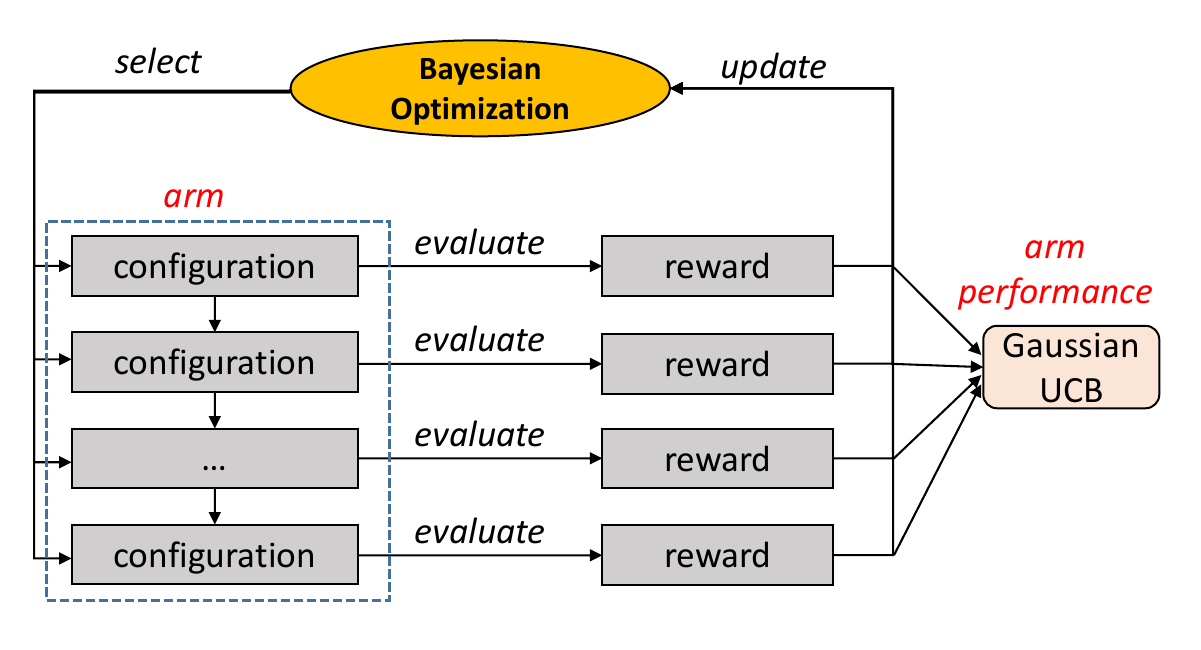}
	\vspace{-2ex}
	\caption{The process of arm evaluation in each round}
	\label{fig:bandit}
\end{figure}
Moreover, it is also noteworthy that the performance of an arm depends on its configuration. 
Different configurations may result in different evaluation results.
We take the model selection as an example.
The hyperparameter configuration plays an important role in the performance of learning models.
Thus, to judge the quality of an arm in each round, it is necessary to evaluate one arm multiple times under the allocated resources.
As shown in Figure~\ref{fig:bandit}, different configurations are used in the process of arm evaluation. 
In this paper, we employ Bayesian optimization to select configurations for each arm, i.e., each arm corresponding to a Bayesian optimization model. 
By learning from previous sampled configurations, the Bayesian optimization method outperforms the random search for selecting good configurations. 
In practice, the evaluation reward can be the prediction accuracy of the learning model. 
According to the evaluation rewards using different configurations, we then compute the Gaussian UCB and view the Gaussian UCB score as the performance metric of the arm.

In this paper, we combine BO with ASF to speed up the AutoML tasks.
Next, we introduce three key steps of BOASF: Bayesian optimization, adaptive arm filtering, and adaptive resource allocation. 

\subsection{Bayesian Optimization} 
We can model the evaluation reward of each arm as a loss function $f:\mathcal{X} \rightarrow \mathbb{R}$ of its configuration $\bm{x} \in \mathcal{X}$.
The configurations can include both discrete and continuous dimensions.
Then, our goal is to find $\bm{x^{\ast}} \in \arg \min_{\bm{x}\in{\mathcal{X}}}f(\bm{x})$.
Bayesian optimization uses a probabilistic model $p(f|D)$ to model the objective function $f$ based on the already observed data points $D=\{ (\bm{x}_0, y_0),\dots, (\bm{x}_{i-1}, y_{i-1}) \}$.
Each data point is a map from the configuration to its evaluation reward.
An acquisition function $a: \mathcal{X}\rightarrow \mathbb{R}$ based on the current model $p(f|D)$ is used to trade off exploration and exploitation.
Typically, BO iterates the following three steps: 
\begin{enumerate*}[label=(\arabic*)]
    \item select a new configuration $\bm{x}_{new} = \arg\max_{\bm{x}\in \mathcal{X}} a(\bm{x})$ that maximizes the acquisition function,
    \item evaluate the configuration and achieve a new data point $(\bm{x}_{new}, y_{new})$,
    \item refit and update the model $p(f|D)$ with $D \cup (\bm{x}_{new}, y_{new})$.
\end{enumerate*}
Expected improvement (EI) is a common acquisition function over the currently best-observed value $y^{\ast} = \min \{ y_0,\dots,y_n \}$:
\begin{equation}\label{equa-ei}
    a(\bm{x}) = \int \max(0, y^{\ast} - f(\bm{x}))dp(f|D)
\end{equation}

Different from the GP-based Bayesian optimization method, the \textbf{Tree Parzen Estimator TPE}~\cite{DBLP:conf/nips/BergstraBBK11} defines $p(x|y)$ using two such densities:

\begin{equation}
    \label{equa-tpe}
    \begin{array}{l}
        \displaystyle l(\bm{x}) = p(y< y^{\ast} | \bm{x}, D) \\[1.5mm]
        \displaystyle g(\bm{x}) = p(y\geq y^{\ast} | \bm{x}, D)
    \end{array}
\end{equation}
Whereas the GP-based approach favored quite an aggressive $y^{\ast}$(typically less than the best observed), the TPE algorithm depends on a $y^{\ast}$  that is larger than the best-observed $f(x)$ so that some points can be used to form  $l(\bm{x})$.
It has been proved that the maximization of ratio ${l(\bm{x})}/{g(\bm{x})}$ is equivalent to maximizing EI in Equation~\eqref{equa-ei}.
Compared to GP-based approach, the TPE scales linearly in the number of data points and can handle both discrete and continuous variables.
In this paper, we use TPE to determine the configurations during the arm evaluation process in each round.

\subsection{Adaptive Arm Filtering}
The filtering criteria of arms depend on the evaluation performance. 
Given a new problem domain, the prior knowledge about which arms can achieve better performance is unknown. 
To address the issue, we allocate equal resources to each arm for pure exploration in the first evaluation round.
After that, as shown in Figure 2, based on the rewards returned within the given resources using different configurations, we estimate the potential of each arm and further determine which arms should be discarded as early as possible. 
To evaluate the potential of each arm, we leverage the Gaussian UCB~\cite{DBLP:journals/jmlr/KaufmannCG12,gaussin_ucb} as the performance metric. 

Gaussian UCB takes the statistical data of historical results into consideration.
Note that each arm is evaluated multiple times in a single evaluation round.
Let $N$ denotes the number of tries.
The formal definition of Gaussian UCB can be expressed as follows:
\begin{equation}\label{eq:ucb}
    \textrm{UCB}(a)=\mu+c * \frac{\sigma}{\sqrt{N}}
\end{equation}
where $\mu$ denotes the mean value of rewards returned by the arm $a$, $\sigma$ denotes the standard deviation. 
The Gaussian UCB can be viewed as the weighted sum of mean and standard deviation, where mean and standard deviation represent exploration and exploitation respectively.
The positive constant $c$ is used to control exploration and exploitation.
%
The high mean value ensures the performance of the arm cannot be too poor. Besides, the standard deviation and the number of tries describe the uncertainty, which also reflects the potential to return an optimal reward.

According to the rewards evaluated with the given resources, we can compute each arm’s Gaussian UCB score and filter some arms with a lower score. 
In contrast to HyperBand~\cite{DBLP:journals/jmlr/LiJDRT17} that directly filters half of the arms, ASF filters the arms adaptively without a fixed ratio. 
Specifically, ASF employs the \emph{MinMaxScaler} to scale the Gaussian UCB score of every arm to a value whose range is between 0 to 1. 
This value is then translated to the probability of advancing to the next evaluation round. 
We can compute the probability using the following equation.
\begin{equation}
    p(a)=\frac{\textrm{UCB}(a)-\min_{a_{i} \in B} {\rm UCB}\left(a_{i}\right)}{\max _{a_{i} \in B} {\rm UCB}\left(a_{i}\right)-\min _{a_{i} \in B} {\rm UCB}\left(a_{i}\right)}
\end{equation}

The \emph{MinMaxScaler} has a good property that the minimum value will be scaled to 0 and the maximum value will be scaled to 1. 
As a result, a high Gaussian UCB score leads to a high probability of advancing.
The arm with the highest Gaussian UCB score is certainly be selected as a candidate arm that advances to the next evaluation round. 
Similarly, the bandit with the worst Gaussian UCB score will be filtered certainly. 
This property ensures that ASF discards at least one arm in a single round of filtering. 
Meanwhile, ASF can retain at least one arm after each round. 

\subsection{Adaptive Resource Allocation}
The adaptive arm filtering strategy causes that an arm with a low Gaussian UCB score still has a probability of advancing to the next evaluation round. 
There may exist significant performance differences between selected arms. 
Thus, it is unfair to allocate equal resources to each arm of the next round.
To tackle the problem, we propose an adaptive resources allocation strategy based on the Gaussian UCB score. 
The arm with a higher Gaussian UCB score will be allocated more resources. 
Moreover, even if a bad arm is not filtered, the resources allocated to it is also fewer.

We leverage the \emph{Softmax} function to allocate resources to each arm.
The input of the \emph{Softmax} function is the Gaussian UCB score. 
The \emph{Softmax} function transforms the Gaussian UCB score of each arm to the proportion of the resources allocated in each evaluation round.
The property of the \emph{Softmax} function ensures that the proportion of the resources is sum to 1.
Let $B$ denote the set of arms and $\overline{R}$ the total available resources in each round of resource allocation. 
Actually, $\overline{R} = R/r$, where $R$ and $r$ represent the total resources and the total rounds respectively.
The resources allocated to each arm $s$ is computed as follows:
\begin{equation}
    res(a)=\frac{e^{{\rm UCB}(a)}}{\sum_{a_{i} \in B} e^{{\rm UCB}\left(a_{i}\right)}} * \overline{R}
\end{equation}

In summary, BOASF consists of two stages: arm filtering and resources allocation. These two stages alternate until there are no resources left. 
Moreover, we use BO to select configurations for each arm.
Algorithm \ref{alg:1} shows the overall process of BOASF.
Finally, we can get the best evaluation result through the historical evaluations.
Also, BOASF is naturally apt to parallel implementation. Take the total CPU time as the total resources $R$ for example. Since each arm is independent, we can evaluate each arm in parallel. Given a multi-core system, different cores can process different arms. Once the allocated resources of each arm are determined, we can evaluate each arm in parallel.

\begin{algorithm}[t]
    \caption{BOASF}
    \label{alg:1}
    \begin{algorithmic}[1]
        \REQUIRE $X_{\rm{train}}, Y_{\rm{train}}$, the total resources $R$, the number of rounds $r$, all arm instances $B$, $c$ in UCB score
        \ENSURE the arm $a^*$ with the best evaluation result

        \STATE divide $R$ into $r$ parts and each part has $\overline{R}$ resources
        \FOR {$i = 1 ... r$}
            \IF {$i == 1$}
                \STATE allocate resources to each arm in $B$ by splitting $\overline{R}$ resources equally
            \ELSE
                \STATE res = Softmax(UCB($B$)) *$\overline{R}$
            \ENDIF
            \FOR {$a \in B$}
            \WHILE{res($a$) $>$ 0}
              \STATE select a configuration \emph{conf} using $\rm{BO_a}$ 
              \STATE evaluate $a$ with \emph{conf} on $X_{\rm{train}}, Y_{\rm{train}}$
              \STATE update $\rm{BO_a}$ model with $\emph{conf}$ and its evaluation result
              \STATE update res($a$) by subtracting the resource cost in the evaluation process  
            \ENDWHILE
                \STATE $\textrm{UCB} (a)=\mu+c * \frac{\sigma}{\sqrt{N}}$
            \ENDFOR
            \STATE $\overline{B} = \{\}$ 
            \FOR {$a \in B$}
                \STATE $\displaystyle p(a)=\frac{{\rm UCB}(a)-\min _{a_{i} \in B} {\rm UCB}\left(a_{i}\right)}{\max _{a_{i} \in B} {\rm UCB}\left(a_{i}\right)-\min _{a_{i} \in B} {\rm UCB}\left(a_{i}\right)}$
                \STATE add $a$ to $\overline{B}$ if $a$ is not filtered
            \ENDFOR
            \STATE $B = \overline{B}$
        \ENDFOR
        \RETURN the bandit $a^*$ with the best evaluation result
\end{algorithmic}
\end{algorithm}

\section{BOASF FOR AUTOML}
The model selection and hyperparameter optimization are two important AutoML problems. 
Each of them can be translated into the $k$ expensive bandit problem. 
Thus, we can tackle these two problems based on the proposed method BOASF.
Next, we introduce how to use BOASF to solve model selection and hyperparameter optimization under a unified multi-armed bandit framework.

\subsection{BOASF for Model Selection}
In the scenario of model selection, BOASF views every candidate machine learning model as an arm.  
The configuration chosen in each evaluation round represents the hyperparameters of the machine learning model.
The total resources can be the available time budget, which is first divided into several equal parts.
Each part represents the available time resources in the corresponding evaluation round. 

In the first round, BOASF allocates the same resources to each machine learning model. 
We use the Bayesian optimization method TPE to tune the hyperparameters of a given model. 
TPE requires less computation time than the GP-based BO approach. 
The number of model instances trained with allocated time resources depends on the time complexity of the model. 
For the model with low time complexity, we can train more instances. 
We evaluate each model instance and view the cross-validation performance as the reward if the model instance can be trained successfully. 
Otherwise, BOASF directly discards the failed models. 
Based on the evaluation results of multiple attempts using different hyperparameters, we compute the Gaussian UCB score of each arm. 
Then, we can perform adaptive arm filtering and resources allocation according to the Gaussian UCB score.

\subsection{BOASF for Hyperparameter Optimization
}
Different from the model selection problem, we do not have a default arm set to run BOASF. 
Given a machine learning model, the search space of hyperparameter optimization may be very huge, especially when the number of hyperparameters is large. 
Thus, it is impossible to view each hyperparameter combination as an arm. 
To address the hyperparameter optimization problem using BOASF, we should first construct an appropriate arm set. 
For this purpose, we design a novel sub-space partitioning method, which divides the hyperparameter search space into several disjoint sub-spaces. 
Each sub-space can be viewed as an arm.

Specifically, we split the range of each hyperparameter into $k$ intervals. 
The combination of intervals from different hyperparameter is regarded as the sub-space. Suppose that the number of hyperparameters is $n$. Then, the total number of sub-spaces is $k^n$. 
For example, given a model with two hyperparameters, one is a continuous hyperparameter whose range is (0, 1), and the other is a discrete hyperparameter whose range is [2,3,4,5]. 
Assume that each hyperparameter is split into two intervals. Then, after sub-space partitioning, there are 4 hyperparameter sub-spaces including: [(0,0,5), [2,3]],[(0,0.5), [4,5]],[[0.5, 1), [2,3]] and [[0.5, 1), [4,5]]. 
Next, we view each hyperparameter sub-space as an arm and employ BOASF for hyperparameter optimization.
When evaluating the performance of an arm, we select hyperparameter configurations from the corresponding hyperparameter sub-space using TPE.
The machine learning model is trained with the selected hyperparameters.
The remaining process is similar to the model selection.

\section{EXPERIMENT}
In this section, we evaluated the performance of BOASF. Moreover, we compared BOASF with AutoSklearn in the model selection task and TPE with original search space in the hyperparameter optimization task.

\subsection{Experiment Settings}
To evaluate the performance of BOASF, we randomly selected 24 datasets from OpenML ~\cite{DBLP:journals/corr/VanschorenRBT14} repository for binary and multi-class classification tasks. 
Each dataset contains no more than 50,000 samples and has no missing values.
Besides, we also selected 6 commonly-used classification datasets from UCI, LibSVM~\cite{DBLP:journals/tist/ChangL11}, and KDDCup.

To reduce the risk of overfitting, we used the 3-fold cross-validation~\cite{DBLP:conf/ijcai/Kohavi95} for model evaluation. 
We chosen the balanced accuracy score as the performance metric, which is also used in AutoSklearn~\cite{DBLP:conf/nips/FeurerKESBH15}.
In the Gaussian UCB equation, we set the positive constant $c$ to 2. 
Moreover, we performed 5 runs for each AutoML methods and analyzed the statistical performance.

Similar to AutoSklearn, we limited the time budget of model training in a single evaluation, which ensures that BOASF and other AutoML methods have enough chances to explore various algorithm models.
We implemented BOASF on the widely-used machine learning package scikit-learn~\cite{DBLP:journals/sigmobile/VaroquauxBLGPM15}.
It offers a wide range of well established and efficiently-implemented ML algorithms.
We ran all experiments on the machines with 12 logical cores in total and 32 GB memory.

\subsection{Model Selection}
In the model selection scenario, we used 16 classification algorithms as the candidate models.
They are AdaBoost, Bernoulli Na\"ive Bayes, Decision Tree, Extra Trees,
Gradient Boost Tree, XGboost, Passive aggressive,
Linear Discriminant Analysis, Quadratic Discriminant Analysis, 
Support Vector Machine, Linear Support Vector Machine,
Multinomial Na\"ive Bayes, Gaussian Na\"ive Bayes,
SGD, Random forest, $k$ Nearest Neighbors respectively.
We compared BOASF with the following three baselines:
\begin{itemize}[itemindent=12pt]
	\item \textbf{SelectBest}. This method trains all classifiers with default hyperparameters respectively and selects the classifier with the best performance. This has been proven to be a competitive baseline~\cite{Ridd:2014:UMP:3015544.3015551}.
    \item \textbf{RandomForest\&TPE}. Random forest is a commonly-used classification algorithm. We used the Bayesian optimization method TPE to tune the hyperparameters.
    \item \textbf{AutoSklearn}. AutoSklearn is the current state-of-the-art approach and is the winner of the ChaLearn AutoML competition. Moreover, it has been shown that AutoSklearn outperforms other AutoML methods in most cases~\cite{Balaji2018BenchmarkingAM}. We turn off the meta learning and include no preprocesssors.
\end{itemize}

For a fair comparison, we turn off the meta learning and include no preprocessors (except 3 preprocessors which are ``balancing'', ``rescaling'' and ``variance\_threshold'' since they cannot be turned off in its implementation) in AutoSklearn. Therefore, we include these preprocessors when do the model selection comparison experiments.
Moreover, the classification algorithms and hyperparameter space in BOASF are exactly the same as those in AutoSklearn.
We also limited the runtime for a single model to 120 seconds.

\begin{table*}
	\centering 
	\caption{Performance comparison between SelectBest, RandomForest\&TPE, AutoSklearn, and BOASF using 30 datasets. Performance is the balanced accuracy score for classification tasks. The bold indicates the best method.} 
	\scalebox{0.76}{ 
		\begin{tabular}[t]{l|cccc|cccr}  
			\toprule 
			\multirow{2}{*}  { Dataset} &\multicolumn{4}{c|}{$R$=2 hours, $r$=3}  &\multicolumn{4}{c}{$R$=4 hours, $r$=3} \\ 
			\cmidrule(lr){2-9} 
			&  SelectBest &RandomForest\&TPE & AutoSklearn  &BOASF   & SelectBest &RandomForest\&TPE & AutoSklearn  & BOASF \\ 
			\midrule 
			adult &0.777 &0.828 &0.831 &\textbf{0.839} &0.777 &0.827 &0.829 &\textbf{0.841}  \\ 
			digits &0.97 &0.963 &0.989 &\textbf{0.992} &0.97 &0.961 &0.992 &\textbf{0.994}  \\ 
			gisette &0.977 &0.962 &\textbf{0.978} &0.975 &0.977 &0.964 &\textbf{0.979} &0.978  \\ 
			wine &0.977 &0.977 &0.993 &0.993 &0.977 &0.972 &0.993 &0.993  \\ 
			kddcup09 &0.528 &0.588 &0.739 &\textbf{0.744} &0.528 &0.586 &0.744 &\textbf{0.745}  \\ 
			letter &\textbf{0.962} &0.943 &0.961 &0.959 &0.962 &0.946 &0.963 &\textbf{0.968}  \\ 
			openml\_8 &0.1 &0.186 &0.181 &\textbf{0.217} &0.1 &0.184 &0.181 &\textbf{0.195}  \\ 
			openml\_37 &0.728 &0.777 &0.763 &\textbf{0.78} &0.728 &0.775 &0.759 &\textbf{0.781}  \\ 
			openml\_276 &0.35 &0.393 &0.463 &\textbf{0.604} &0.35 &0.402 &0.444 &\textbf{0.575}  \\ 
			openml\_278 &0.807 &0.802 &0.807 &\textbf{0.832} &0.807 &0.802 &0.836 &\textbf{0.856}  \\ 
			openml\_279 &0.98 &0.98 &0.98 &\textbf{0.992} &0.98 &0.984 &0.99 &\textbf{0.992}  \\ 
			openml\_285 &0.44 &0.487 &0.494 &\textbf{0.578} &0.44 &0.488 &0.529 &\textbf{0.634}  \\ 
			openml\_337 &0.81 &0.867 &0.845 &\textbf{0.886} &0.81 &0.867 &0.844 &\textbf{0.9}  \\ 
			openml\_459 &0.792 &0.816 &0.843 &\textbf{0.862} &0.792 &0.818 &0.855 &\textbf{0.862}  \\ 
			openml\_475 &0.424 &0.404 &0.435 &\textbf{0.47} &0.424 &0.413 &0.445 &\textbf{0.456}  \\ 
			openml\_683 &0.797 &0.771 &0.816 &\textbf{0.851} &0.797 &0.79 &0.836 &0.836  \\ 
			openml\_714 &0.585 &0.57 &0.657 &0.657 &0.585 &0.593 &0.656 &\textbf{0.691}  \\ 
			openml\_724 &0.831 &0.847 &\textbf{0.867} &0.856 &0.831 &0.853 &0.856 &\textbf{0.858}  \\ 
			openml\_726 &0.82 &0.826 &\textbf{0.864} &0.835 &0.82 &0.819 &\textbf{0.9} &0.853  \\ 
			openml\_731 &0.711 &0.776 &0.775 &\textbf{0.777} &0.711 &0.762 &0.775 &\textbf{0.789}  \\ 
			openml\_733 &0.933 &0.895 &0.957 &\textbf{0.961} &0.933 &0.913 &0.957 &\textbf{0.961} \\ 
			openml\_736 &0.747 &0.775 &\textbf{0.795} &0.783 &0.747 &0.775 &0.773 &\textbf{0.794}  \\ 
			openml\_750 &0.634 &\textbf{0.657} &0.63 &0.654 &0.634 &\textbf{0.677} &0.633 &0.672  \\ 
			openml\_753 &0.619 &0.635 &0.699 &\textbf{0.703} &0.619 &0.632 &0.716 &\textbf{0.749}  \\ 
			openml\_763 &0.845 &0.845 &\textbf{0.9} &0.885 &0.845 &0.84 &\textbf{0.91} &0.875  \\ 
			openml\_764 &0.805 &0.887 &0.889 &\textbf{0.897} &0.805 &0.879 &0.89 &\textbf{0.892}  \\ 
			openml\_771 &0.743 &0.757 &0.789 &\textbf{0.795} &0.743 &0.737 &0.789 &\textbf{0.795}  \\ 
			openml\_773 &0.835 &0.855 &0.855 &0.855 &0.835 &0.85 &0.86 &0.86  \\ 
			openml\_774 &0.568 &0.546 &0.557 &\textbf{0.597} &0.568 &0.553 &0.571 &\textbf{0.604}  \\ 
			openml\_783 &0.742 &0.738 &\textbf{0.847} &0.772 &0.742 &0.722 &\textbf{0.867} &0.794  \\ 
			\bottomrule 
		\end{tabular} 
	}
	\label{tb:exp-results-2-4-hour-all-info-without-CR-2} 
\end{table*}

We compared BOASF with three baselines on a wide range of datasets (e.g., 30 datasets in total).
For BOASF, the total number of evaluation rounds $r$ is 3. 
We also set the total time resource $R$ to be 2 and 4 hours respectively. 
Table~\ref{tb:exp-results-2-4-hour-all-info-without-CR-2} shows the results of  performance comparison.
From Table~\ref{tb:exp-results-2-4-hour-all-info-without-CR-2}, we can see that no method can achieve the best performance on all datasets.
BOASF outperforms the other three methods in most cases.
Since BOASF tunes the hyperparameters of each arm with Bayesian optimization in each evaluation round,  it performs better than SelectBest that uses the default hyperparameters.
Although we perform hyperparameter optimization for the random forest model,  the performance of RandomFrest\&TPE is still lower than AutoSklearn and BOASF.
This is because both AutoSklearn and BOASF can automatically select the best model.

As shown in Table~\ref{tb:exp-results-2-4-hour-all-info-without-CR-2}, BOASF outperforms AutoSklearn under different time limits.
Specifically, in the case of 2 hours, BOASF performs better than AutoSklearn on 20/30 datasets, ties it on 3 datasets.
For the time limit of 4 hours, the performance of BOASF is superior to AutoSklearn on 23/30 datasets and the same as AutoSklearn on 3 datasets.
Thus, in the model selection scenario, BOASF can achieve stable performance improvement on different budgets and shows robust anytime performance.

%

%
AutoSklearn views the choice of the algorithm itself as a hyperparameter and employs the Bayesian optimization to solve the high-dimensional hyperparameter optimization problem.
Without an appropriate algorithm, the performance improvement with hyperparameter optimization is limited.
In contrast, BOASF views the model selection as a finite multi-armed bandit problem.
Then, it filters the poor-performed models as quickly and allocates more resources to more promising models adaptively.
Moreover, BO is used to optimize the hyperparameters of each model.
Thus, BOASF can find high-quality models in most cases and has robust anytime performance under different time limits.

\subsection{Hyperparameter Optimization}
%
%
In the hyperparameter optimization scenario, we compared BOASF with TPE which works on the entire search space to demenstrate the effectiveness of dividing the hyperparammeter search space into several disjoint sub-spaces. 
Typically, we use the implementation of TPE in Hyperopt~\cite{bergstra2015hyperopt}. We denote this TPE setting as Hyperopt-full.   
We also compared BOASF with BOHB~\cite{DBLP:conf/icml/FalknerKH18} in the same experimental setting. 

BOHB is the state-of-the-art hyperparameter optimization method that outperforms random search~\cite{DBLP:journals/jmlr/BergstraB12} and HyperBand~\cite{DBLP:journals/jmlr/LiJDRT17}.
We compared these three methods on two widely-used models: random forest and logistic regression.
Similar to model selection, different time limits ($2$ h and $4$ h) are used to evaluate hyperparameter optimization.
The hyperparameters in BOASF and BOHB are $c=2, r=3$ and $b_{min}=\frac{1}{27}, b_{max}=1, \eta=3$ respectively.
The random forest model is configured with 5 hyperparameters:
\begin{itemize}[itemindent=12pt]
	\item \textbf{splitting criterion}, ``gini" or ``entropy".
	\item \textbf{max features}, uniform float in $[0.5, 1]$.
	\item \textbf{min samples split}, uniform integer in $[2,21]$.
	\item \textbf{min samples leaf}, uniform integer in $[1,21]$.
	\item \textbf{bootstrap}, ``true" or ``false".
\end{itemize} 
The logistic regression model contains 3 hyperparameters:
\begin{itemize}[itemindent=12pt]
	\item \textbf{penalty}, ``L1" or ``L2".
	\item \textbf{C}, uniform float in $[1\mathrm{e}{-4}, 1\mathrm{e}{4}]$.
	\item \textbf{max iteration}, uniform integer in $[50,500]$.
\end{itemize}

We divided the range of each hyperparameter into $2$ intervals. 
Therefore, BOASF has $2^5$ and $2^3$ arms respectively.
Table~\ref{tb:exp-results-2-4-hour-hyp-lr} and Table~\ref{tb:exp-results-2-4-hour-hyp-rf}
 shows that BOASF achieved better balanced accuracy score than Hyperopt-full and BOHB on both the logistic regression and random forest settings statistically.
%
%
%
The experimental results between BOASF and Hyperopt-full demonstrate that BOASF along with the sub-space partitioning strategy is effective for hyperparameter optimization.

\subsection{Internal Framework Study}
We additionally performed experimentation to study the influences of the two hyperparameters in BOASF: $c$ of Gaussian UCB in Equation~\ref{eq:ucb} and the total number of evaluation rounds $r$. 
Take BOASF on hyperparameter optimization on logistic regression for example, we do a grid search on $c=[2,3,4,5]$ and $r=[2,3,4,5]$. The total time budget of BOASF and Hyperopt-full are both 1 hour.

\begin{figure}[!ht]
	\centering
	\scalebox{1.0}{
	\begin{subfigure}[b]{0.5\columnwidth}
		\centering
		\includegraphics[width=1.0\columnwidth]{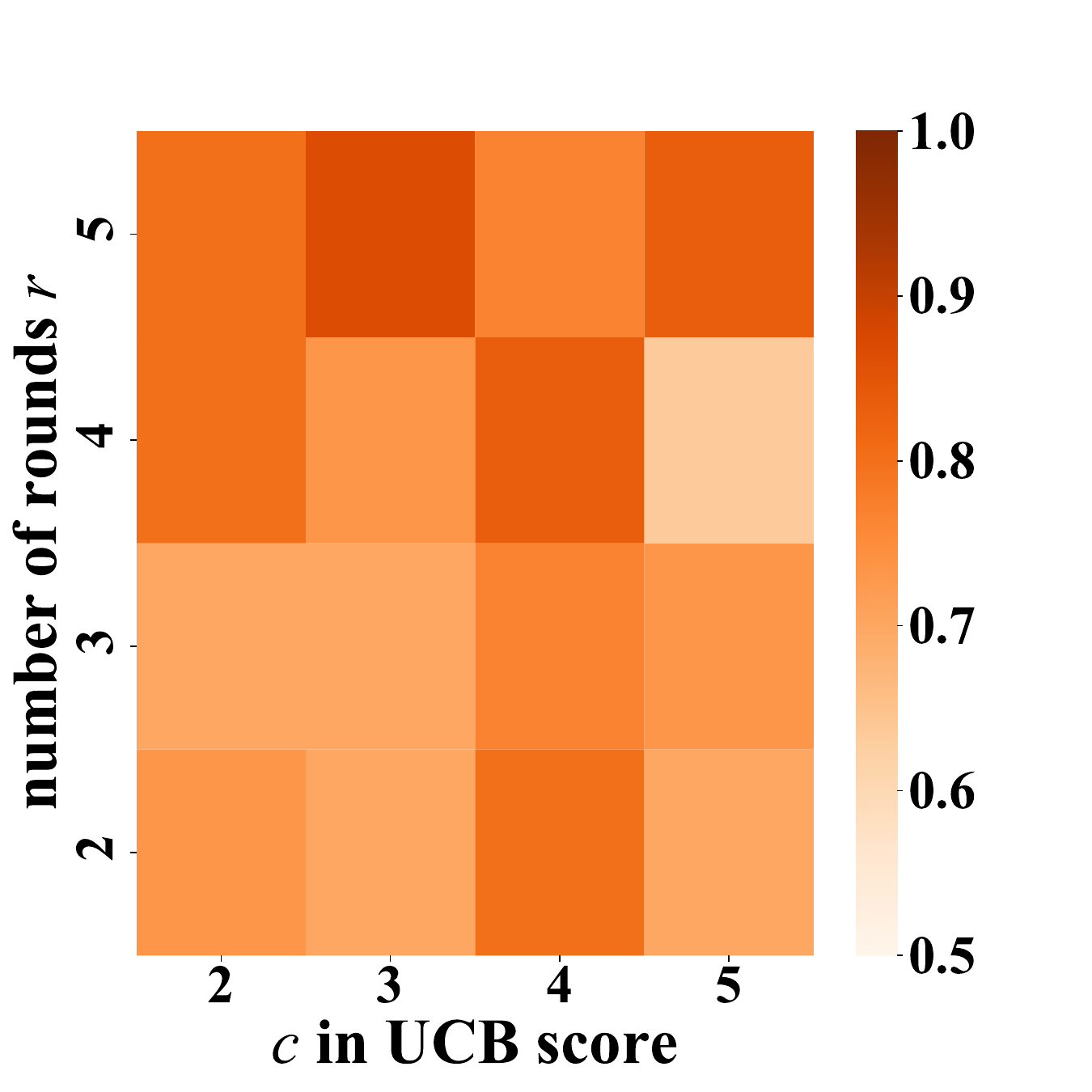}
		\caption[]{{\small 3-fold cross validation score}}
	\end{subfigure}
	\hfill
	\begin{subfigure}[b]{0.5\columnwidth}
		\centering
		\includegraphics[width=1.0\columnwidth]{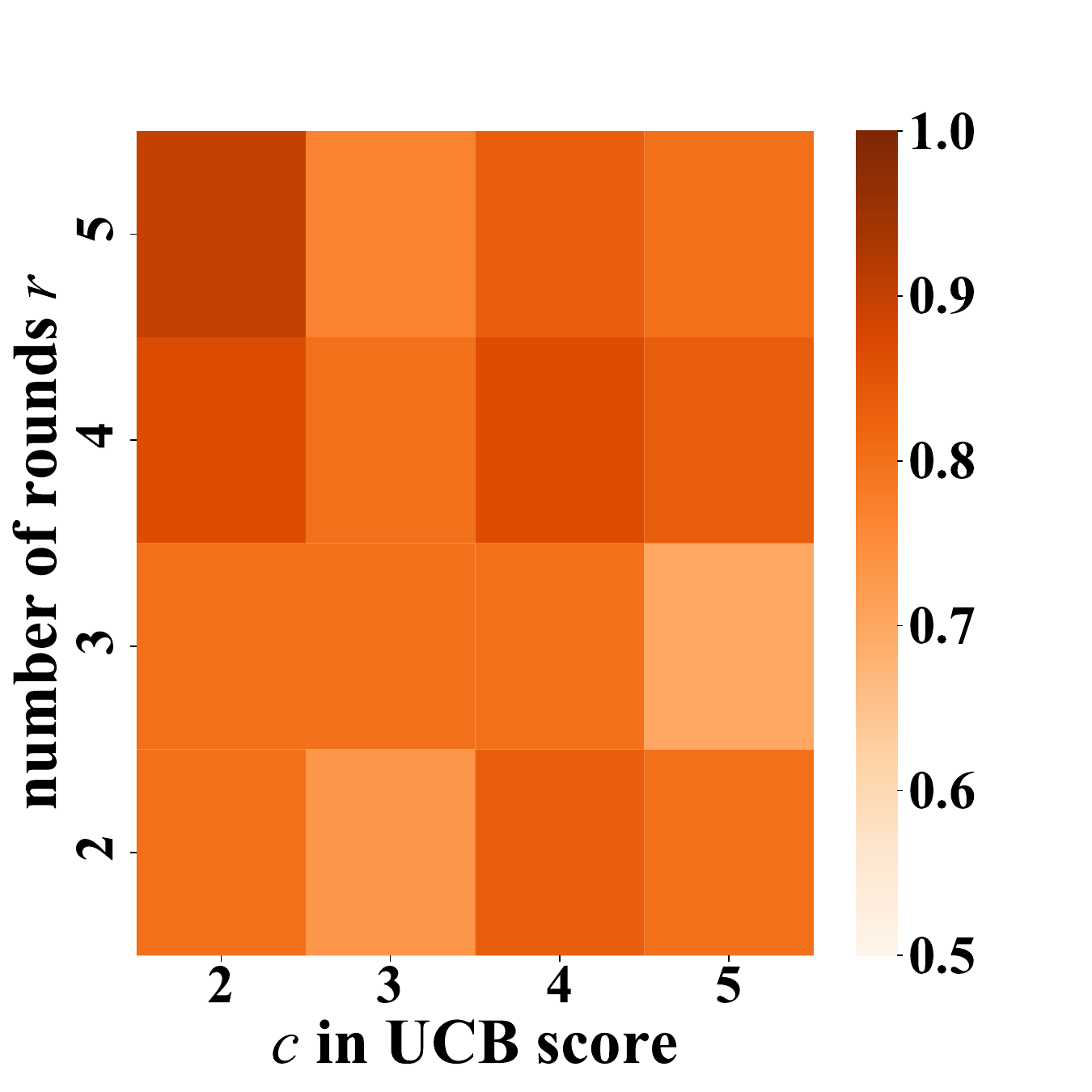}
		\caption[]{{\small Test score}}
	\end{subfigure}
	}
	\caption{The ratio of datasets which BOASF outperformed or tied with Hyperopt-full under different combinations of hyperparameters ($c$ and $r$) in BOASF.}
	\label{fig:ratio}
\end{figure}

Figure~\ref{fig:ratio} shows the ratio of 30 datasets that BOASF outperforms Hyperopt-full.
BOASF achieved robust performance against differernt $c$ and $r$ settings on both cross validation score and test score.

\begin{table}
    \centering 
    \caption{Performance comparison between Hyperopt-full, BOHB and BOASF for hyperparameter optimization with Logistic Regression (3-fold cross validation balanced accuracy score).}
	\scalebox{0.72}{ 
		\begin{tabular}{l|ccc|ccr}
			\toprule
			\multirow{2}{*}{Dataset} &\multicolumn{3}{c|}{$2$ hours} &\multicolumn{3}{c}{$4$ hours} \\
			\cmidrule(lr){2-7} 
			&\multicolumn{1}{c}{Hyperopt-full} &\multicolumn{1}{c}{BOHB} &\multicolumn{1}{c|}{BOASF} &\multicolumn{1}{c}{Hyperopt-full} &\multicolumn{1}{c}{BOHB} &\multicolumn{1}{c}{BOASF} \\
			\midrule 
			adult & 0.660 & \textbf{0.766} & 0.766 & \textbf{0.662} & 0.660 & 0.660 \\
			digits & \textbf{0.953} & 0.938 & 0.950 & \textbf{0.954} & 0.950 & 0.954 \\
			gisette & 0.972 & 0.972 & \textbf{0.973} & 0.972 & 0.972 & \textbf{0.973} \\
			wine & 0.941 & 0.927 & \textbf{0.986} & 0.941 & \textbf{0.944} & \textbf{0.944} \\
			kddcup09 & 0.503 & \textbf{0.503} & 0.502 & 0.502 & 0.501 & \textbf{0.503} \\
			letter & \textbf{0.769} & 0.722 & 0.723 & \textbf{0.770} & 0.769 & 0.769 \\
			openml\_8 & 0.070 & 0.061 & \textbf{0.105} & 0.071 & \textbf{0.677} & 0.070 \\
			openml\_37 & \textbf{0.732} & 0.715 & 0.726 & \textbf{0.731} & 0.718 & 0.731 \\
			openml\_276 & \textbf{0.344} & 0.303 & 0.326 & 0.266 & 0.233 & \textbf{0.358} \\
			openml\_278 & \textbf{0.814} & 0.778 & 0.795 & 0.808 & \textbf{0.808} & \textbf{0.808} \\
			openml\_279 & 0.097 & 0.088 & \textbf{0.823} & \textbf{0.098} & 0.092 & 0.097 \\
			openml\_285 & 0.360 & 0.312 & \textbf{0.373} & 0.369 & \textbf{0.452} & 0.364 \\
			openml\_337 & 0.767 & 0.780 & \textbf{0.786} & 0.758 & \textbf{0.768} & 0.765 \\
			openml\_459 & 0.825 & \textbf{0.828} & 0.811 & 0.825 & \textbf{0.828} & 0.825 \\
			openml\_475 & \textbf{0.403} & 0.389 & 0.398 & 0.403 & 0.364 & \textbf{0.418} \\
			openml\_683 & 0.687 & \textbf{0.692} & 0.668 & 0.687 & \textbf{0.692} & 0.687 \\
			openml\_714 & 0.602 & \textbf{0.602} & \textbf{0.602} & 0.602 & \textbf{0.602} & \textbf{0.602} \\
			openml\_724 & \textbf{0.511} & 0.504 & 0.504 & \textbf{0.511} & 0.507 & 0.511 \\
			openml\_726 & 0.665 & 0.666 & \textbf{0.745} & 0.665 & \textbf{0.665} & \textbf{0.665} \\
			openml\_731 & 0.709 & 0.700 & \textbf{0.747} & 0.710 & 0.670 & \textbf{0.710} \\
			openml\_733 & \textbf{0.935} & 0.924 & 0.924 & \textbf{0.935} & 0.906 & 0.935 \\
			openml\_736 & 0.725 & 0.712 & \textbf{0.736} & \textbf{0.725} & 0.725 & 0.725 \\
			openml\_750 & 0.512 & 0.492 & \textbf{0.520} & 0.515 & 0.472 & \textbf{0.517} \\
			openml\_753 & \textbf{0.670} & 0.664 & 0.658 & \textbf{0.677} & 0.663 & 0.676 \\
			openml\_763 & 0.795 & 0.747 & \textbf{0.795} & 0.795 & \textbf{0.799} & 0.795 \\
			openml\_764 & 0.500 & 0.500 & 0.500 & 0.500 & 0.500 & 0.500 \\
			openml\_771 & 0.733 & 0.729 & \textbf{0.743} & 0.733 & 0.729 & \textbf{0.733} \\
			openml\_773 & 0.815 & \textbf{0.825} & 0.820 & 0.815 & 0.790 & \textbf{0.815} \\
			openml\_774 & 0.568 & \textbf{0.586} & \textbf{0.586} & 0.568 & \textbf{0.568} & \textbf{0.568} \\
			openml\_783 & 0.589 & 0.498 & \textbf{0.616} & \textbf{0.589} & 0.481 & 0.589 \\
			\bottomrule 
		\end{tabular}
	}
	\label{tb:exp-results-2-4-hour-hyp-lr}
\end{table}

\begin{table}
    \centering 
    \caption{Performance comparison between Hyperopt-full, BOHB and BOASF for hyperparameter optimization on Random Forest (3-fold cross validation balanced accuracy score).}
	\scalebox{0.72}{ 
		\begin{tabular}{l|ccc|ccr}
			\toprule
			\multirow{2}{*}{Dataset} &\multicolumn{3}{c|}{$2$ hours} &\multicolumn{3}{c}{$4$ hours} \\
			\cmidrule(lr){2-7} 
			&\multicolumn{1}{c}{Hyperopt-full} &\multicolumn{1}{c}{BOHB} &\multicolumn{1}{c|}{BOASF} &\multicolumn{1}{c}{Hyperopt-full} &\multicolumn{1}{c}{BOHB} &\multicolumn{1}{c}{BOASF} \\
			\midrule 
			adult & 0.781 & \textbf{0.783} & 0.781 & 0.782 & 0.781 & \textbf{0.827} \\
			digits & \textbf{0.962} & 0.959 & 0.959 & \textbf{0.964} & 0.960 & 0.957 \\
			gisette & 0.385 & 0.490 & \textbf{0.492} & 0.475 & 0.495 & \textbf{0.505} \\
			wine & 0.972 & \textbf{0.979} & 0.970 & 0.979 & \textbf{0.979} & 0.977 \\
			kddcup09 & 0.503 & 0.498 & \textbf{0.510} & 0.500 & 0.505 & \textbf{0.528} \\
			letter & \textbf{0.950} & 0.944 & 0.950 & \textbf{0.951} & 0.950 & 0.941 \\
			openml\_8 & 0.077 & \textbf{0.743} & 0.083 & 0.078 & 0.082 & \textbf{0.186} \\
			openml\_37 & 0.755 & 0.761 & \textbf{0.765} & 0.762 & 0.762 & \textbf{0.775} \\
			openml\_276 & 0.365 & 0.289 & \textbf{0.392} & \textbf{0.386} & 0.289 & 0.386 \\
			openml\_278 & \textbf{0.802} & 0.802 & 0.802 & \textbf{0.802} & 0.802 & 0.796 \\
			openml\_279 & 0.000 & \textbf{0.093} & 0.000 & 0.000 & 0.090 & \textbf{0.802} \\
			openml\_285 & 0.455 & 0.381 & \textbf{0.489} & 0.458 & 0.429 & \textbf{0.511} \\
			openml\_337 & \textbf{0.848} & 0.843 & 0.841 & 0.840 & 0.847 & \textbf{0.867} \\
			openml\_459 & 0.795 & \textbf{0.830} & 0.799 & 0.818 & 0.830 & \textbf{0.836} \\
			openml\_475 & \textbf{0.405} & 0.374 & 0.404 & \textbf{0.413} & 0.403 & 0.402 \\
			openml\_683 & 0.767 & 0.718 & \textbf{0.790} & 0.790 & 0.718 & \textbf{0.790} \\
			openml\_714 & 0.565 & 0.513 & \textbf{0.579} & 0.562 & 0.500 & \textbf{0.573} \\
			openml\_724 & 0.842 & \textbf{0.847} & 0.844 & 0.839 & 0.850 & \textbf{0.857} \\
			openml\_726 & 0.826 & 0.826 & \textbf{0.837} & 0.826 & 0.826 & \textbf{0.827} \\
			openml\_731 & 0.762 & 0.631 & \textbf{0.762} & 0.762 & 0.640 & \textbf{0.777} \\
			openml\_733 & 0.910 & 0.884 & \textbf{0.919} & \textbf{0.915} & 0.880 & 0.906 \\
			openml\_736 & 0.691 & 0.602 & \textbf{0.693} & 0.691 & 0.635 & \textbf{0.775} \\
			openml\_750 & 0.659 & 0.552 & \textbf{0.662} & 0.649 & 0.628 & \textbf{0.660} \\
			openml\_753 & 0.627 & 0.602 & \textbf{0.637} & 0.624 & 0.605 & \textbf{0.634} \\
			openml\_763 & 0.850 & 0.819 & \textbf{0.856} & \textbf{0.865} & 0.827 & 0.850 \\
			openml\_764 & 0.782 & 0.763 & \textbf{0.798} & 0.782 & 0.763 & \textbf{0.886} \\
			openml\_771 & 0.757 & \textbf{0.771} & 0.757 & 0.741 & \textbf{0.761} & 0.751 \\
			openml\_773 & 0.830 & \textbf{0.845} & \textbf{0.845} & 0.845 & \textbf{0.855} & 0.850 \\
			openml\_774 & 0.546 & \textbf{0.565} & 0.544 & 0.541 & \textbf{0.573} & 0.555 \\
			openml\_783 & \textbf{0.753} & 0.560 & 0.738 & \textbf{0.732} & 0.460 & 0.718 \\
			\bottomrule 
		\end{tabular}
	}
	\label{tb:exp-results-2-4-hour-hyp-rf}
\end{table}

\section{RELATED WORK}
\subsection{Hyperparameter Optimizaion}
Hyperparameter optimization has received more and more attention from both academia and industry. Previous research work shows that random search method outperforms the widely-used grid search method~\cite{DBLP:journals/jmlr/BergstraB12}. 
Recently, the model-based Bayesian optimization method has emerged as a popular hyperparameter optimization method, which aims to find the global optimal value from a black-box function~\cite{DBLP:journals/corr/abs-1012-2599}. The commonly-used Bayesian optimization methods includes GP~\cite{DBLP:conf/nips/SnoekLA12}, SMAC~\cite{DBLP:conf/lion/HutterHL11}, TPE~\cite{DBLP:conf/nips/BergstraBBK11}. 
Besides the Bayesian optimization method, a multi-armed bandit based method Hyperband~\cite{DBLP:journals/jmlr/LiJDRT17} was proposed. Hyperband is able to discard the poor-performed hyperparameters as early as possible using the successive halving method. Recently, BOHB that combines Hyperband and Bayesian optimization was proposed to achieve better performance~\cite{DBLP:conf/icml/FalknerKH18,DBLP:journals/corr/abs-1801-01596}.

\subsection{Model Selection}
To achieve model selection and hyperparameter optimization jointly, the researchers viewed the AutoML problem as a combined algorithm selection and hyperparameter optimization(CASH) problem.
Recently, two practical AutoML system Auto-WEKA~\cite{DBLP:conf/kdd/ThorntonHHL13} and AutoSklearn~\cite{DBLP:conf/nips/FeurerKESBH15} were design to address the CASH problem using the tree-based Bayesian optimization method SMAC. Compared to Auto-WEKA, AutoSklearn integrates the meta-learning and the ensemble leaning to improve the efficiency of AutoML.
The tree-based pipeline optimization (TPOT) method~\cite{DBLP:conf/gecco/OlsonBUM16} was proposed to automatically design and optimize machine learning pipelines using genetic programming.
TPOT uses the idea from the ensemble method stacking to construct new features. Therefore, TPOT is more time-consuming than other AutoML methods.

\subsection{Multi-Armed Bandit Problem}
The multi-armed bandit (MAB) is a problem in which a fixed limited set of resources must be allocated between competing arms in a way that maximizes their expected gain~\cite{Lai:1985:AEA:2609660.2609757}.
Each arm provides a random reward from a probability distribution specific to that arm. 
The objective of the bandit is to maximize the sum of rewards earned through a sequence of pulls. 
The key of MAB is the trade-off between exploration and exploitation.
Many algorithms such as $\epsilon$-greedy~\cite{Sutton:1998:IRL:551283}, UCB~\cite{Auer:2002:FAM:599614.599677} have been proposed to solve the MAB problem.
However, these algorithms focus on maximizing
the cumulative rewards over time~\cite{DBLP:journals/corr/abs-1204-5721}.

In the context of AutoML, we need to solve the best-arm identification (or pure exploration) MAB problem.
Pure exploration bandit problems aim to minimize the simple regret as quickly as possible in any given setting~\cite{bubeck2009pure,Carpentier,AAAI1714705}.
Existing work addresses the best arm objective in the stochastic setting.
Recently, the non-stochastic algorithms have been proposed for hyperparameter optimization~\cite{Jamieson2015NonstochasticBA,DBLP:journals/jmlr/LiJDRT17}.
In this paper, we focus on the non-stochastic pure exploration MAB problem under a limited resource budget.
Note that the cost of evaluating an arm can be drastically more than pulling it.
We propose adaptive successive filtering and resource allocation to speed up both model selection and hyperparameter optimization under a unified MAB framework.

\section{DISCUSSION AND CONCLUSION}
In this paper, we proposed an easy-to-implement and effective AutoML method called BOASF that combines Bayesian optimization with adaptive successive filtering to speed up AutoML under a unified multi-armed bandit framework. 
BOASF not only can discard the poor-performed arms as quickly, but also adaptively allocate more resources for each promising arms. 
Moreover, the configurations of each arm are selected using BO. 
BOASF can be easily extended to support both model selection and hyperparameter optimization. 
For the model selection, each available model is viewed as a arm. 
For the hyperparameter optimization, we proposed a novel sub-space partitioning method to view each disjoint hyperparameter sub-space as an arm.
Experimental results on a broad range of datasets show that BOASF statistically outperforms the state-of-the-art AutoML methods in most cases. 
Also, BOASF achieves robust and better anytime performance under different time limits.

There are several improvements to do in our method in the future. Firstly, the UCB score works for a random variable range from $0$ to $1$. Those metrics which do not meet this condition need to be readjusted to this range. Secondly, we use \emph{MinMaxScaler} to transfer the UCB score to filtering probability since it guarantees a property that the lowest score will be discarded and the highest score will advance to the next round. Other rescaling methods which has this property still work in our method. Thirdly and finally, th TPE method is used to tune the hyperparameters of each arm. We can compare it with other Bayesian Optimizaiton methods.

\newpage
\bibliography{reference}
\end{document}